\documentclass[conference]{IEEEtran}
\IEEEoverridecommandlockouts
\usepackage{cite}
\usepackage{amsmath,amssymb,amsfonts}
\usepackage{algorithmic}
\usepackage{graphicx}
\usepackage{textcomp}
\usepackage{xcolor}
\usepackage{booktabs}
\usepackage{multirow}
\usepackage{stfloats}  % 提供跨栏表格的位置控制
\usepackage{float}     % 提供 H 选项
\usepackage{array}
\usepackage{hyperref} % 加载 hyperref 包以生成可点击的链接
\usepackage{adjustbox}
\usepackage{makecell} % 引入 makecell 宏包
\def\BibTeX{{\rm B\kern-.05em{\sc i\kern-.025em b}\kern-.08em
    T\kern-.1667em\lower.7ex\hbox{E}\kern-.125emX}}
\begin{document}

\title{LG-CD: Enhancing Language-Guided Change Detection through SAM2 Adaptation}

\author{
    Yixiao Liu$^{1}$ \quad
    Yizhou Yang$^{1}$ \quad
    Jinwen Li$^{2}$\\
    Jun Tao$^{1}$\quad
    Ruoyu Li$^1$ \quad
    Xiangkun Wang$^1$ \quad
    Min Zhu$^{1,*}$ \quad
    Junlong Cheng$^{1,*}$ \\
    $^1$College of Computer Science, Sichuan University, China  \\
    $^2$School of Computer Science and Technology, Xinjiang University, China  \\
     $^*$ Corresponding Author \\
     \small \{liuyixiao1, yang\_yizhou\}@stu.scu.edu.cn, 107556522204@stu.xju.edu.cn, \\
     \{2022141460135, 2022141460358\}@stu.scu.edu.cn, \{wangxiangkun, zhumin\}@scu.edu.cn, cjl951015@stu.scu.edu.cn\\
}

\maketitle

\begin{abstract}
Remote Sensing Change Detection (RSCD) typically identifies changes in land cover or surface conditions by analyzing multi-temporal images. Currently, most deep learning-based methods primarily focus on learning unimodal visual information, while neglecting the rich semantic information provided by multimodal data such as text. To address this limitation, we propose a novel \textbf{L}anguage-\textbf{G}uided \textbf{C}hange \textbf{D}etection model (LG-CD). This model leverages natural language prompts to direct the network's attention to regions of interest, significantly improving the accuracy and robustness of change detection. Specifically, LG-CD utilizes a visual foundational model (SAM2) as a feature extractor to capture multi-scale pyramid features from high-resolution to low-resolution across bi-temporal remote sensing images. Subsequently, multi-layer adapters are employed to fine-tune the model for downstream tasks, ensuring its effectiveness in remote sensing change detection. Additionally, we design a Text Fusion Attention Module (TFAM) to align visual and textual information, enabling the model to focus on target change regions using text prompts. Finally, a Vision-Semantic Fusion Decoder (V-SFD) is implemented, which deeply integrates visual and semantic information through a cross-attention mechanism to produce highly accurate change detection masks. Our experiments on three datasets—LEVIR-CD, WHU-CD, and SYSU-CD—demonstrate that LG-CD consistently outperforms state-of-the-art change detection methods. Furthermore, our approach provides new insights into achieving generalized change detection by leveraging multimodal information.
\end{abstract}


\section{Introduction}
\label{sec:intro}

%\IEEEPARstart{C}{hange}
The RSCD task aims to detect changes in surface objects or environmental conditions by analyzing and processing remote sensing images of the same area acquired in different periods. This task is widely used in many fields such as urban planning~\cite{urbanplanning}, disaster assessment~\cite{disasterassessment}, video surveillance~\cite{visualsurveillance}, and natural resource management~\cite{naturalresourcesmanagement1} and has important practical significance.  Traditional change detection methods mainly rely on techniques such as thresholding~\cite{thresholding}, morphological analysis~\cite{morphological}, and image algebra~\cite{imagealgebra}. However, these methods are sensitive to interference factors such as seasonal changes, changes in lighting conditions, and shadows, resulting in significant performance degradation.

In recent years, change detection methods based on deep learning have gradually become mainstream. For example, methods based on convolutional neural networks (CNNs)~\cite{snunet, FC} can more effectively focus on significantly changed areas in images by constructing multi-scale features and introducing attention mechanisms. These methods significantly improve the accuracy of change detection, but the local modeling characteristics of CNNs limit its ability to capture long-range contextual information in remote sensing images. In order to solve this problem, researchers began to introduce the Vision Transformer model~\cite{VIT} to better handle large-scale remote sensing images and complex change patterns. For example, BIT~\cite{BIT} combines CNN and Transformer models to obtain local and global feature information; ChangeFormer~\cite{changeformer} achieves fine-grained change detection through hierarchical self-attention encoder and lightweight decoder.

% Although the above methods have made significant progress, due to the scarcity of remote sensing data, high annotation costs, and the complexity of multi-source data fusion, the visual foundation model (foundation model) performs poorly due to its generalization and transfer capabilities across data sets. The superiority of the remote sensing image change detection task is demonstrated. For example, FastSAM authors and others use FastSAM's pre-trained decoder to extract robust visual features from dual-temporal remote sensing images, and fine-tune it through multi-layer convolution adapters to adapt to the change detection task; SCD-SAM authors and others propose a A context-sensitive semantic change-aware dual encoder that combines MobileSAM with CNN for semantic change detection. However, the above methods mainly focus on single-modal visual information and fail to fully utilize the rich semantic information contained in multi-modal data. This single-modal method limits the generalization ability of the model to a certain extent, making it difficult to cope with complex dynamic practical application scenarios.

Despite the significant advancements achieved by the aforementioned methods, challenges such as the scarcity of remote sensing data, high annotation costs, and the complexity of multi-source data fusion remain unresolved~\cite{review}. Foundation models, known for their strong generalization and transferability across datasets, have demonstrated superior performance in remote sensing change detection tasks. For instance,   Ding et al.~\cite{adaptSAM} utilized the pre-trained encoder of FastSAM~\cite{FASTSAM} to extract robust visual features from bi-temporal remote sensing images, fine-tuning the model for change detection tasks through multi-layer convolutional adapters. Similarly, Mei et al.~\cite{SCD-SAM} proposed a context-sensitive semantic change-aware dual encoder that integrates MobileSAM~\cite{MOBILESAM} and CNN for semantic change detection. However, these methods primarily focus on visual information and fail to fully exploit the rich semantic information embedded in multimodal data. This reliance on unimodal approaches inherently limits the generalization capacity of the models, making them less effective in addressing the complexities and dynamics of real-world application scenarios.

Recent studies have demonstrated that multimodal perception plays a critical role in enhancing the performance of tasks based on a single modality. Liu et al.~\cite{LEVIRCC} were the first to introduce natural language descriptions into change detection tasks, constructing a multimodal change detection dataset.~\cite{CDCHAT} and~\cite{CHANGECLIP} utilized low-rank adaptation (LoRA)~\cite{LORA} fine-tuning and CLIP-based~\cite{CLIP} textual prompts to develop visual-language multimodal change detection (CD) models. These studies indicate that combining language and visual modalities not only provides additional contextual information for the model but also significantly improves the accuracy and robustness of change detection.

This paper proposes a new visual-linguistic multi-modal learning method to solve the bottleneck of change detection in remote sensing images. Specifically, we utilize the pre-trained SAM2 encoder~\cite{SAM2} as the shared visual feature extractor and CLIP (Contrastive Language-Image Pretraining)~\cite{CLIP} as the text feature extractor. Then, a text fusion attention module is designed to align text with visual features. Furthermore, we build a visual-semantic fusion decoder to generate highly accurate change detection masks. Experimental results show that combining visual features with textual semantic information not only improves the model's semantic understanding capabilities, but also significantly improves detection accuracy and generalization capabilities. By leveraging the contextual information provided by natural language, the model can better understand complex scenes and changing patterns, thereby achieving robust detection of multiple types of targets. The main contributions of this paper are summarized as follows:

\textbf{1.} We introduce a novel \textbf{L}anguage-\textbf{G}uided \textbf{C}hange \textbf{D}etection model (LG-CD) that seamlessly integrates visual and textual information. By utilizing natural language cues to direct attention toward target regions, the model achieves substantial improvements in detection accuracy and robustness.

\textbf{2.} The model makes full use of SAM2's high generalization and portability across different data sets, and designs multi-layer adapters for fine-tuning to ensure high adaptability in change detection tasks and enhance multi-scale semantic understanding.

\textbf{3.} We design a text fusion attention module and a visual-semantic fusion decoder. The text fusion attention module weights visual features according to text cues and focuses on key changing areas. The fused decoder integrates visual and semantic information through cross-attention to generate high-precision change masks.

\begin{figure*}[!t]
	\centering
	\includegraphics[width=6.5in]{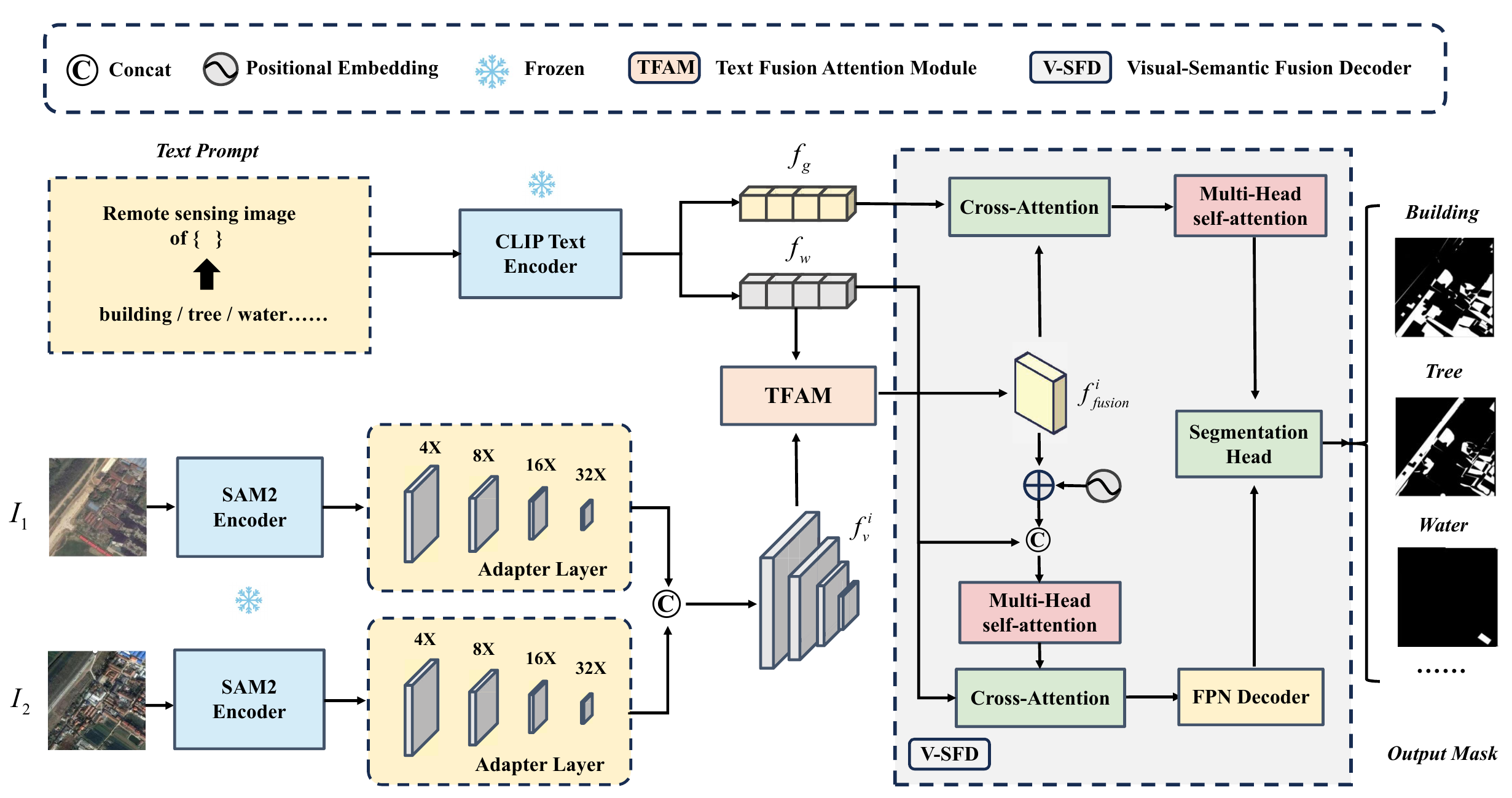}
	\caption{Our LG-CD pipeline accepts two remote sensing images captured at different time points, along with their corresponding text prompts, as inputs. The Adapter Layer is utilized to adapt to change detection tasks, TFAM integrates text features into visual features, and V-SFD deeply fuses visual and semantic information to generate highly accurate change detection masks.}
	\label{fig1}
\end{figure*}

\section{METHOD}
\subsection{Overview}
The overall structure of LG-CD is shown in Fig \ref{fig1}. First, the SAM2 encoder extracts multi-scale features from the bi-temporal images and is fine-tuned through a multi-level adapter. A text attention module is employed to align textual and image features, guiding the model to focus on the detection regions. Subsequently, the visual-semantic fusion decoder integrates both visual and semantic information to generate the final change detection mask. 

\subsection{SAM2 Encoder and Adapters}
LG-CD receives two remote sensing image data of different phases $I_{1}, I_{2} \in \mathbb{R}^{H \times W \times 3}$. First, the SAM2 encoder is used to extract multi-scale features from these two phase images: $f^i_{1}, f^i_{2} \in \mathbb{R}^{\frac{H}{2^{(i+2)}} \times \frac{W}{2^{(i+2)}} \times C_{i}}$. SAM2 uses Hiera image encoder for image encoding. The encoder is a hierarchical visual Transformer architecture that applies windowed absolute position embedding and interpolated global position embedding and uses a feature pyramid network to fuse features at different stages, where $i=0,1,2,3$, generating feature maps downsampled by 4 times, 8 times, 16 times, and 32 times, respectively. $C_0$ to $C_3$ represent the number of channels at different scales.

To capture task-relevant multi-scale information, we introduce lightweight adapter layers to fine-tune the output features of the encoder. These adapters consist of convolutional layers, each applied to the multi-scale features of the two temporal images. Subsequently, we concatenate the feature maps output by the adapters along the channel dimension to generate a fused global feature map. The specific process can be expressed as:
\begin{align}
	f_v^{i}=Adapter(f^i_1) \text { © } Adapter(f^i_2)
\end{align}

Here, ``$\text{©}$" represents the channel concatenation operation, and $Adapter$ refers to the combination of a 1×1 convolution, Batch Normalization, and ReLU activation function. The multi-level adapter design ensures that features at each level can be independently refined and optimized, laying a strong foundation for the seamless integration of multi-scale textual information.

\begin{figure}[!t]
	\centering
	\includegraphics[width=1\linewidth]{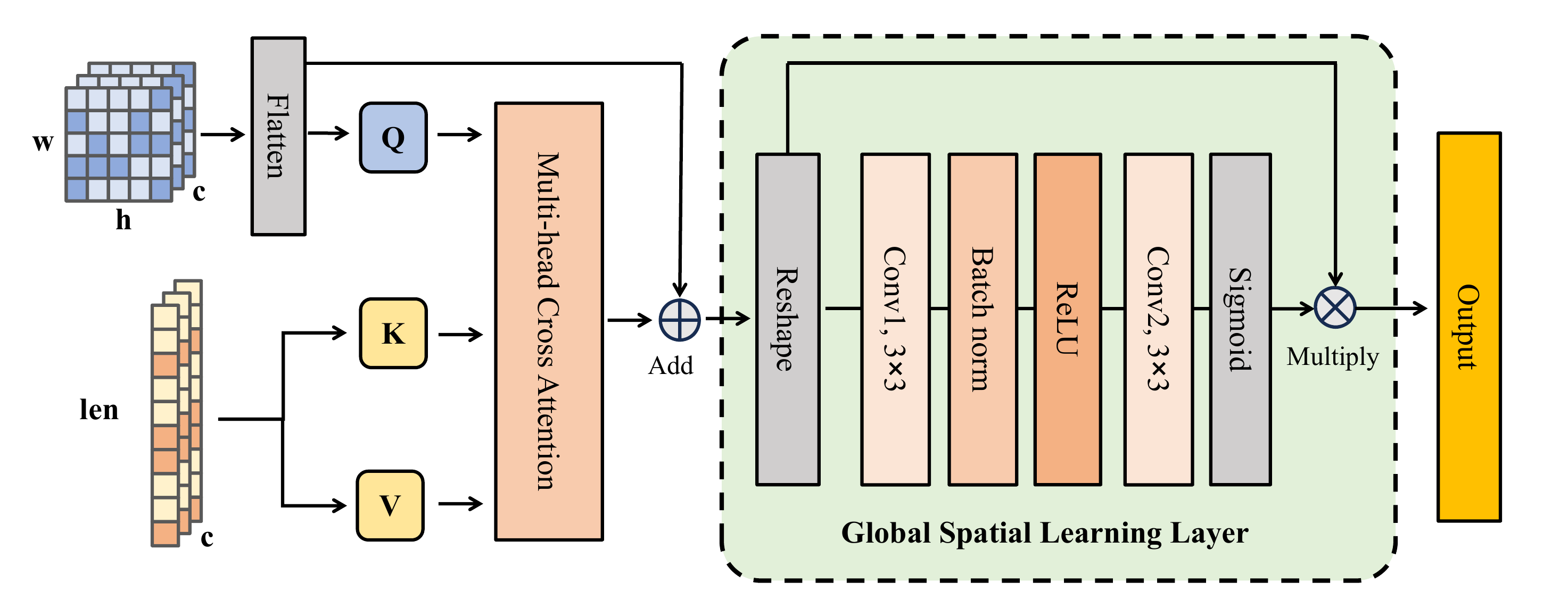}
	\caption{Structure of the proposed TFAM.}
	\label{fig2}
\end{figure}

\subsection{Text Fusion Attention Module}

The Text Fusion Attention Module (TFAM) is designed to effectively integrate textual prompt features into visual features, providing the model with clear task direction and attention focus. Specifically, for the input textual prompt $T$, we first utilize the CLIP~\cite{CLIP} model to encode it, obtaining semantic embeddings as follows:
\begin{align}
    f_{\text {w }}, f_{\text {g }}= CLIP_{text}(T),
  \end{align}
  
where $f_{\text{w}}$ represents the word-level embedding, which captures fine-grained semantics and contextual information for each word in the text;$f_{\text{g}}$ is the global text embedding, which characterizes the overall meaning and intent of the entire sentence.

As shown in Fig \ref{fig2}, we treat multi-scale visual features as queries and word embedding features as key-value pairs. Through a multi-head cross-attention (MCA) mechanism, we extract task-relevant semantic information from the word embedding features, which is then fused back into the visual features. This process is formulated as:
\begin{align}
    \widehat{f_{v}} &= \operatorname{MCA}\left(f_{v}^{i}, f_{w}\right) \nonumber \\
    &= \operatorname{softmax} \left(\frac{W_{q}\left(f_{v}^{i}\right)^{T} W_{k}\left(f_{w}\right)}{\sqrt{C^{i}}}\right) W_{v}\left(f_{w}\right)^{T},
    \label{eq:mca}
  \end{align}
  
where $W_{q}$, $W_{k}$, $W_{v}$ are linear transformation functions that project the input features into the query, key, and value subspaces, respectively; $C^{i}$  is the number of channels in the $i$-th feature map.

To further enhance the model's spatial awareness, we introduce a Global Spatial Learning Layer (Fig \ref{fig2}). This layer first generates a spatial attention feature map through convolution operations, highlighting the most relevant spatial regions. Subsequently, the spatial attention feature map is element-wise multiplied with the fused visual features, integrating global spatial context into the feature representation and producing the final fused features $f_{fusion}^i$. 

% This process can be formalized as:
% \begin{align}
%   f_{fusion}^{i}=\widehat{f_{v}} * \operatorname{sigmoid}\left\{\operatorname{Conv}_{2}\left\{\gamma\left\{B N\left[\operatorname{Conv}_{1}\left(\widehat{f_{v}}\right)\right]\right\}\right\}\right\}, \label{eq:SA}
% \end{align}

% where $\operatorname{Conv_{1}}$ and $\operatorname{Conv_{2}}$ are convolutional layers with a kernel size of 3x3, $BN$ denotes batch normalization, and $\gamma$ is a ReLU activation function. Through this series of operations, the model is better able to capture the interaction between global and local information, thus improving segmentation accuracy and robustness in complex scenarios.

\subsection{Visual-Semantic Fusion Decoder}
The visual-semantic fusion decoder is the core component of LG-CD, which is used to further integrate multi-modal information based on the multi-scale features $f_{fusion}$, word-level embedding features $f_w$, and global text embedding features $f_g$, so as to generate language-guided accurate change detection results. Specifically, for the fusion features $f_{fusion}^{i}$ of the $i$th scale, we first flatten them and add positional encoding to preserve the spatial position information. Then, the fusion features with positional encoding are directly fused with the word embedding features to form multimodal tags. Finally, multi-head self-attention (MSA) and multi-head cross-attention (MCA) are used to extract the relevant information between them, so as to capture the intra-modal and inter-modal dependencies. Note that MCA takes visual features as queries, word embedding features as keys and values, only outputs visual tags for subsequent processes, and discards word embedding tags. This process can be formalized as: 
\begin{align}
&f_{fusion}^{i} = Flatten\left(f_{fusion}^{i}\right) + \operatorname{Pos}_{sin}, \\
&f_{MSA}^{i} = \operatorname{MSA}\left(f_{fusion}^{i} \text{ © } f_{w}\right), \\
&f_{MCA}^{i} = \operatorname{MCA}\left(f_{MSA}^{i}, f_{w}\right) \label{eq:MCA}, \\
&f_{V} = \operatorname{FPN}\left(f_{MCA}^{i}\right),
\end{align}

where $\operatorname{Pos}_{sin}$ is the sinusoidal position embedding. Finally, a structure similar to FPN~\cite{FPN} is used to integrate the aligned visual features $f_V$ of all scales.

\begin{table}[t]
\centering
\caption{Details of datasets used in the experiment.}
\begin{minipage}{0.45\textwidth}
\centering
\begin{tabular}{l| l| l| l}
\toprule
\textbf{Dataset} & \textbf{Category} & \textbf{Split (Train/Val/Test)} & \textbf{Total Samples} \\ 
\midrule
LEVIR-CD & Building & 445 / 64 / 128 & 637 \\
WHU-CD & Building & 5948 / 743 / 743 & 7434 \\
SYSU-CD & Multi-class & 10000 / 0 / 2000 & 12000 \\
\bottomrule
\end{tabular}
\end{minipage}
\label{table1}
\end{table}

In order to make full use of the global language instructions to adjust the visual features, we still use the attention operation to achieve this goal, but the steps are adjusted. Specifically, we first use the global text embedding feature $f_{g}$ as the query of the cross-attention, and $f_{fusion}^{i}$ as the key and value, so as to integrate the global semantic information into the visual features. Subsequently, self-attention is used to further integrate the initial language cues with the semantic features adjusted by content awareness to achieve the organic fusion of multimodal information: 
\begin{align}
f_{L}=\operatorname{MSA}\left(\operatorname{MCA}\left(f_{\text {g }}, f_{\text {V }}\right)\right),
\end{align}

\begin{table*}[ht]
\centering
\setlength{\tabcolsep}{5pt} % 减小列间距离
\renewcommand{\arraystretch}{1.2} % 调整行高以提高可读性
\caption{Quantitative results of different CD methods obtained on the Levir-CD, WHU-CD and SYSU-CD datasets.\\ Color convention: \textcolor{red}{\textbf{best}}, \textcolor{blue}{\textbf{2nd-best}}. All values are given in percentages (\%).}
\begin{adjustbox}{width=\textwidth} % 强制表格宽度等于文本宽度
\begin{tabular}{lccccc|ccccc|ccccc}
\toprule
\multirow{2}{*}{\textbf{Method}} & \multicolumn{5}{c|}{\textbf{LEVIR-CD}} & \multicolumn{5}{c|}{\textbf{WHU-CD}} & \multicolumn{5}{c}{\textbf{SYSU-CD}} \\
\cmidrule{2-16}
& \textbf{$Pre$} & \textbf{$Rec$} & \textbf{$IoU$} & \textbf{$F_1$} & \textbf{$OA$} & \textbf{$Pre$} & \textbf{$Rec$} & \textbf{$IoU$} & \textbf{$F_1$} & \textbf{$OA$} & \textbf{$Pre$} & \textbf{$Rec$} & \textbf{$IoU$} & \textbf{$F_1$} & \textbf{$OA$} \\
\midrule
FC-EF~\cite{FC}        & 87.93  & 82.77 & 77.55 & 85.18 & 98.39 & 84.08  & 77.45 &72.54  &80.49  &97.86 & 76.60 & 76.21 & 63.18 & 75.81 & 88.32 \\
FC-Siam-Conc~\cite{FC}  &91.11 	&81.91 	&78.52 	&86.22 	&98.55 	&62.07 	&83.19 	&65.31 	&65.73 	&97.10 & 79.72 & 77.02 & 66.08 & 77.73 & 89.24 \\
FC-Siam-Diff~\cite{FC}  &91.00 	&82.52 	&78.36 	&86.42 	&98.53 	&61.15 	&86.08 	&60.56 	&70.07 	&96.52 &\textcolor{blue}{\textbf{84.85}}  & 65.14 & 59.08 & 71.84 & 85.62 \\
SNUNet~\cite{snunet}   &90.62 	&\textcolor{blue}{\textbf{88.31}} 	&80.88 	&89.32 	&\textcolor{blue}{\textbf{98.82}} 	&87.53 	&84.09 	&75.40 	&85.62 	&98.65 & 79.64 & 77.39 & 67.42 & 78.45 &\textcolor{blue}{\textbf{90.11}}  \\
BIT~\cite{BIT}     &90.13 	&88.12 	&81.80 	&89.01 	&98.72 	&87.37 	&86.80 	&79.17 	&87.01 &\textcolor{blue}{\textbf{98.72}} & 82.18 & 75.53 & 66.61 & 78.66 & 89.75 \\
ChangeFormer~\cite{changeformer} &\textcolor{blue}{\textbf{91.37}} 	&88.01 	&\textcolor{red}{\textbf{83.39}} 	&\textcolor{blue}{\textbf{89.80}} 	&98.77 	&\textcolor{blue}{\textbf{91.06}} 	&\textcolor{blue}{\textbf{89.75}} 	&\textcolor{blue}{\textbf{86.12}} &\textcolor{blue}{\textbf{90.36}} 	&99.12 & 82.32 &\textcolor{blue}{\textbf{77.59}}  &\textcolor{blue}{\textbf{66.81}}  &\textcolor{blue}{\textbf{79.85}}  & 89.99 \\
\midrule
LG-CD (proposed) &\textcolor{red}{\textbf{91.51}} &\textcolor{red}{\textbf{89.96}} &\textcolor{blue}{\textbf{83.36}} 	&\textcolor{red}{\textbf{90.35}}  &\textcolor{red}{\textbf{99.13}} &\textcolor{red}{\textbf{92.31}} &\textcolor{red}{\textbf{91.75}} &\textcolor{red}{\textbf{90.47}} &\textcolor{red}{\textbf{91.83}} &\textcolor{red}{\textbf{99.51}} & \textcolor{red}{\textbf{84.88}} & \textcolor{red}{\textbf{80.38}} & \textcolor{red}{\textbf{70.59}} & \textcolor{red}{\textbf{80.48}} & \textcolor{red}{\textbf{91.84}}
\\
\bottomrule
\end{tabular}\label{tab1}
\end{adjustbox}
\end{table*}

After obtaining the integrated visual features $f_V$ and the content-aware language embedding $f_L$, we input them into the segmentation head for similarity calculation (matrix multiplication) to generate a response map. Finally, the response map is upsampled by bilinear interpolation, and the final output segmentation mask is obtained by threshold (binarization) operation.

\section{EXPERIMENTS}

\subsection{Implementation Details}

The training of LG-CD was completed on an NVIDIA RTX 3090 GPU (24GB). The encoder was initialized with pre-trained SAM2 weights, and during training, the encoder parameters were kept frozen. To enhance the model's generalization ability, data augmentation techniques such as sliding window cropping, random cropping, and random flipping were applied to the input data. The Adam optimizer was used during the optimization process, with an initial learning rate set to 0.0001 and a batch size of 4. For the loss function, a combination of cross-entropy loss ($L_{CE}$), IoU loss ($L_{IoU}$)~\cite{iou}, and Dice loss ($L_{Dice}$)~\cite{dice} was used to minimize the difference between the predicted mask $Y_p^{i}$ and the ground truth $Y_t$:
\begin{equation}
  \begin{split}
  L_{total} &= \frac{1}{n} \sum_{i=1}^{n}\Bigg[
  \left(1-\alpha-\beta \right)L_{CE}\left(Y_{p}^{i}, Y_t\right) \\
  &\quad + \alpha \times L_{IoU}\left(Y_{p}^{i}, Y_t\right) \\
  &\quad + \beta \times L_{Dice}\left(Y_{p}^{i}, Y_t\right)
  \Bigg],
  \end{split}
  \end{equation}

\begin{figure*}[!t]
	\centering
	\includegraphics[width=7in]{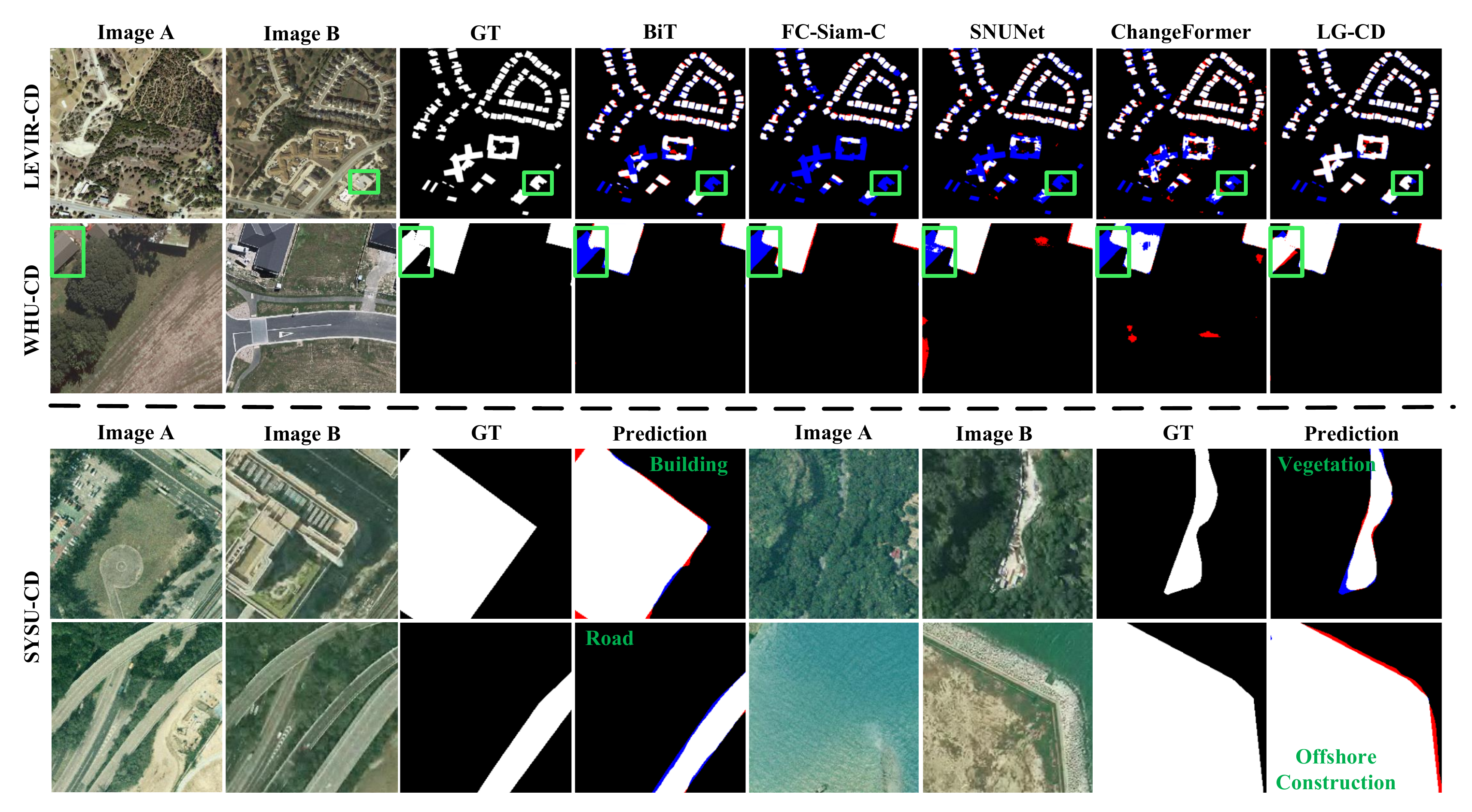}
	\caption{The upper two rows present a qualitative comparison of different methods on the LEVIR-CD and WHU-CD datasets. The lower two rows display change detection results on the LG-CD dataset under various text prompts. In the figure, white, black, red, and blue represent true positives, true negatives, false positives, and false negatives, respectively.}
	\label{fig3}
\end{figure*}

Our model outputs six predicted probability maps by default, i.e., $n=6$.The weights of the three losses are controlled by the parameters $\alpha$ and $\beta$. Through empirical experiments, we set $\alpha$ and $\beta$ to 0.2 and 0.1, respectively, to achieve a good balance between performance and stability. Finally, we use Precision ($Pre$), Recall ($Rec$), F1-score ($F_1$), Intersection over Union ($IoU$), and Overall Accuracy ($OA$) to evaluate the performance of all comparison methods~\cite{BIT,changeformer}. In this study, we utilized three widely recognized remote sensing change detection datasets: LEVIR-CD~\cite{LEVIR}, WHU-CD~\cite{WHU}, and SYSU-CD~\cite{SYSUCD}.
Table \ref{table1} provides a detailed overview of the dataset's categories and subdivisions. SYSU-CD is a multi-class dataset encompassing four categories: Building, Road, Vegetation, and Offshore Construction. All comparative models were evaluated using identical dataset splits to ensure fair and reliable results.

\subsection{Comparison with State-of-the-Art Methods}

As shown in Table \ref{tab1}, we conducted a comprehensive performance comparison of the proposed LG-CD method with six other state-of-the-art remote sensing change detection methods~\cite{FC,snunet,BIT,changeformer} across three datasets. The experimental results demonstrate that LG-CD consistently achieves either the best or second-best performance on different datasets, highlighting its strong adaptability and outstanding effectiveness in change detection tasks. Notably, in terms of the critical recall metric, LG-CD outperforms the second-best model by 1.65\%, 2\%, and 2.79\% on the three datasets, respectively. These results not only confirm the robustness and superiority of LG-CD in diverse change detection tasks, but also demonstrate that effectively integrating multimodal information can further improve performance, highlighting the innovation and practical value of our approach.

\subsection{Visual Analysis}

\begin{figure}[!b]
	\centering
	\includegraphics[width=\linewidth]{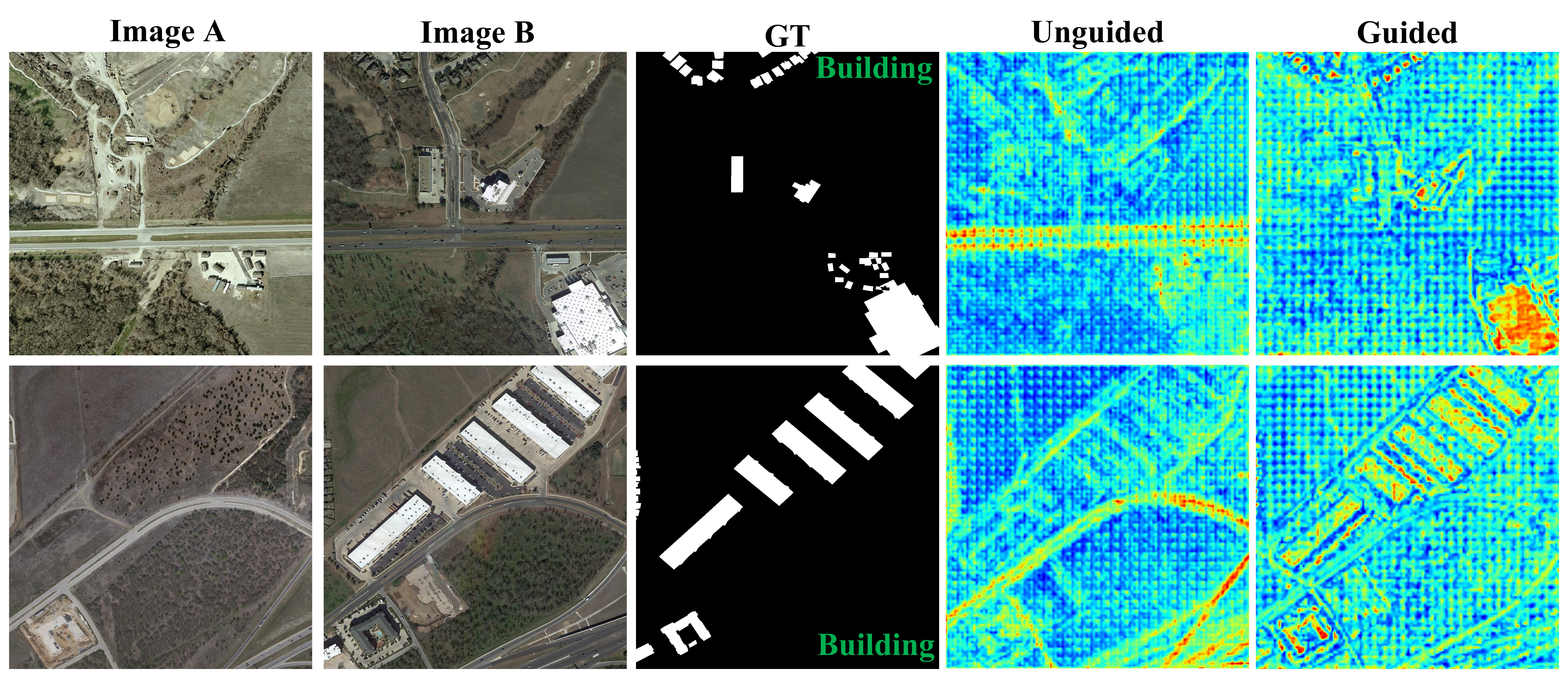}
	\caption{Heatmap visualization of comparison results under text guidance.}
	\label{fig4}
\end{figure}

\begin{table*}[h]
\centering
\caption{Ablation results  on the Levir-CD and WHU-CD datasets. All values are given in percentages (\%).}
\begin{adjustbox}{width=\textwidth} % 强制表格宽度等于文本宽度
\begin{tabular}{lccccc|ccccc}
\toprule
\multirow{2}{*}{\textbf{Method}} & \multicolumn{5}{c|}{\textbf{LEVIR-CD}} & \multicolumn{5}{c}{\textbf{WHU-CD}} \\
\cmidrule{2-11}
& \textbf{$Acc$} & \textbf{$F_1$} & \textbf{$IoU$} & \textbf{$Pre$} & \textbf{$Rec$} & \textbf{$Acc$} & \textbf{$F_1$} & \textbf{$IoU$} & \textbf{$Pre$} & \textbf{$Rec$} \\
\midrule
ResNet + FPN        & 98.89  & 81.22 & 70.65 & 81.13 & 80.84 &95.14  & 72.83 &65.40  &72.42  &77.31 \\
Hiera image encoder + FPN  &99.02 	&84.91 	&74.36 	&86.95 	&84.39 	&98.63 	&81.72 	&71.28 	&82.95 	&80.39 \\
Hiera image encoder + TFAM + FPN  &99.04 	&86.17 	&78.49 	&87.45 	&87.28 	&98.95 	&84.56 	&73.89 	&83.58 	&86.72  \\
Hiera image encoder + V-LFD   &99.10 	&89.15 	&81.32 	&90.02 	&88.31	&99.40 	&90.35 	&88.93 	&91.15 	&91.66 \\
\midrule
  Hiera image encoder + TFAM + V-LFD     &99.13 	&90.35 	&83.36 	&91.51 	&89.96	&99.51 	&91.83 	&90.47 	&92.31 &91.75 \\
\bottomrule
\end{tabular}\label{tab2}
\end{adjustbox}
\end{table*}

As shown in Fig \ref{fig3}, the first two rows present the qualitative experimental results of our method compared with six other state-of-the-art methods. It can be visually observed that LG-CD exhibits fewer blue and red regions in change detection, indicating significantly lower false negatives and false positives. Notably, in the small target areas marked by green boxes, our method demonstrates superior detection performance. The last two rows show that, guided by different language prompts, our method achieves change detection results for different entities. This further validates the application potential of the proposed method, suggesting that it can be extended to more diverse prompt conditions in the future to achieve universal remote sensing change detection.

We conducted a visual analysis to highlight the importance of incorporating semantic information into visual feature extraction. As shown in Fig \ref{fig4}, without the integration of semantic information, the model simultaneously focuses on multiple targets such as buildings, roads, and bridges. However, after incorporating semantic guidance, the model's attention is concentrated on the specified building targets. This observation demonstrates that embedding semantic information into the visual feature extraction process can effectively focus the model's attention, enabling it to concentrate more on task-relevant features, thereby enhancing both detection performance and task relevance.

% \begin{table}[!htbp]
% \centering 
% \caption{Heatmap visualization results of different methods}
% \setlength{\tabcolsep}{1pt} % 调整列之间的间距
% \renewcommand{\arraystretch}{1.2} % 调整行高
% \begin{tabular}{@{}c
%                 >{\centering\arraybackslash}m{1.5cm} 
%                 >{\centering\arraybackslash}m{1.5cm} 
%                 >{\centering\arraybackslash}m{1.5cm} 
%                 >{\centering\arraybackslash}m{1.5cm} 
%                 >{\centering\arraybackslash}m{1.5cm} 
%                 >{\centering\arraybackslash}m{1.5cm} 
%                 >{\centering\arraybackslash}m{1.5cm}
%                @{}}
    
% % 第一行：标题
% & \textbf{ImageA} & \textbf{ImageB} & \textbf{GT} & \textbf{guided} & \textbf{unguided}\\

% % 1 行
% \textbf{Pair1}&
% \includegraphics[width=1.5cm]{figure/exp_vis/pair7-A.png} &
% \includegraphics[width=1.5cm]{figure/exp_vis/pair7-B.png} &
% \includegraphics[width=1.5cm]{figure/exp_vis/pair7-LABEL.png} &
% \includegraphics[width=1.5cm]{figure/exp_vis/pair7-guided.png} &
% \includegraphics[width=1.5cm]{figure/exp_vis/pair7-unguided.png} &
%  \\

% % 2 行
% \textbf{Pair2}&
% \includegraphics[width=1.5cm]{figure/exp_vis/pair8-A.png} &
% \includegraphics[width=1.5cm]{figure/exp_vis/pair8-B.png} &
% \includegraphics[width=1.5cm]{figure/exp_vis/pair8-LABEL.png} &
% \includegraphics[width=1.5cm]{figure/exp_vis/pair8-guided.png} &
% \includegraphics[width=1.5cm]{figure/exp_vis/pair8-unguided.png} &\\

% \end{tabular}
% \end{table}

\subsection{Ablation Studies}

% As shown in Table 1, we conducted ablation experiments on the LEVIR-CD dataset to assess the impact of incorporating semantic information on model performance, and to verify the effectiveness of the key modules in the model, including the Hiera image encoder, Text Attention Module (TFAM), and Visual-Semantic Fusion Decoder (V-LFD). We selected ResNet as the baseline model and replaced V-LFD with FPN to create a unimodal model, providing a performance benchmark. Then, we incrementally added the TFAM and V-LFD modules.
% As shown in the table, the performance of the baseline model was not very satisfactory. However, when SAM2 was introduced, there was an improvement of 1.69 in F1 and 1.71 in IoU compared to the baseline. After incorporating the V-LFD module, there was a significant improvement of 7.13 in F1 and 8.67 in IoU over the baseline. Finally, the method with all modules achieved the best performance. The continuous improvement in model performance with the introduction of each module demonstrates the contribution of each component and highlights the effective role of semantic information in guiding the model's attention to the change regions, thereby improving detection accuracy.

We conducted ablation experiments on the LEVIR-CD and WHU-CD datasets. As shown in Table \ref{tab2}, using Hiera as the image encoder significantly outperforms ResNet-based encoders~\cite{resnet}. Subsequently, incremental additions of the TFAM module and the V-LFD module both resulted in notable performance improvements. These results not only validate the effectiveness of each module in LG-CD but also demonstrate that incorporating textual information can effectively guide the model to focus on change regions, thereby significantly enhancing change detection accuracy.

\section{Conclusion}

This paper proposes an innovative language-guided change detection model (LG-CD), which integrates visual and semantic information to significantly enhance the accuracy and robustness of remote sensing change detection using natural language prompts. LG-CD leverages the powerful vision foundation model (SAM2) as its backbone network and employs multi-layer adapters for fine-tuning, enabling rapid adaptation to remote sensing change detection tasks. Additionally, we designed attention-based modules, TFAM and V-LFD, to align and deeply fuse visual and language features, thereby effectively capturing change patterns and accurately generating change detection masks. Experimental results on three change detection datasets fully validate the effectiveness of LG-CD, demonstrating its superior performance compared to other methods. In future research, we plan to further extend this framework to accommodate more diverse semantic scenarios, achieving generalized language-guided change detection.

\bibliographystyle{IEEEbib}
\bibliography{main}

\begin{thebibliography}{10}

\bibitem{urbanplanning}
Jorge Prendes, Marie Chabert, Frédéric Pascal, Alain Giros, and Jean-Yves Tourneret,
\newblock ``A new multivariate statistical model for change detection in images acquired by homogeneous and heterogeneous sensors,''
\newblock {\em IEEE Transactions on Image Processing}, vol. 24, no. 3, pp. 799--812, 2015.

\bibitem{disasterassessment}
Zhuo Zheng, Yanfei Zhong, Junjue Wang, Ailong Ma, and Liangpei Zhang,
\newblock ``Building damage assessment for rapid disaster response with a deep object-based semantic change detection framework: From natural disasters to man-made disasters,''
\newblock {\em Remote Sensing of Environment}, vol. 265, pp. 112636, 2021.

\bibitem{visualsurveillance}
Mark~J Carlotto,
\newblock ``Detection and analysis of change in remotely sensed imagery with application to wide area surveillance,''
\newblock {\em IEEE Transactions on image processing}, vol. 6, no. 1, pp. 189--202, 1997.

\bibitem{naturalresourcesmanagement1}
Baudouin Descl{\'e}e, Patrick Bogaert, and Pierre Defourny,
\newblock ``Forest change detection by statistical object-based method,''
\newblock {\em Remote sensing of environment}, vol. 102, no. 1-2, pp. 1--11, 2006.

\bibitem{thresholding}
Qiqi Zhu, Xi~Guo, et~al.,
\newblock ``Land-use/land-cover change detection based on a siamese global learning framework for high spatial resolution remote sensing imagery,''
\newblock {\em ISPRS Journal of Photogrammetry and Remote Sensing}, vol. 184, pp. 63--78, 2022.

\bibitem{morphological}
Aisha Javed, Sejung Jung, Won~Hee Lee, and Youkyung Han,
\newblock ``Object-based building change detection by fusing pixel-level change detection results generated from morphological building index,''
\newblock {\em Remote Sensing}, vol. 12, no. 18, pp. 2952, 2020.

\bibitem{imagealgebra}
Yufei Yang, Jiahui Qu, Song Xiao, Wenqian Dong, Yunsong Li, and Qian Du,
\newblock ``A deep multiscale pyramid network enhanced with spatial--spectral residual attention for hyperspectral image change detection,''
\newblock {\em IEEE Transactions on Geoscience and Remote Sensing}, vol. 60, pp. 1--13, 2022.

\bibitem{snunet}
Sheng Fang, Kaiyu Li, Jinyuan Shao, and Zhe Li,
\newblock ``Snunet-cd: A densely connected siamese network for change detection of vhr images,''
\newblock {\em IEEE Geoscience and Remote Sensing Letters}, vol. 19, pp. 1--5, 2021.

\bibitem{FC}
Rodrigo~Caye Daudt, Bertr Le~Saux, and Alexandre Boulch,
\newblock ``Fully convolutional siamese networks for change detection,''
\newblock in {\em 2018 25th IEEE international conference on image processing (ICIP)}. IEEE, 2018, pp. 4063--4067.

\bibitem{VIT}
Alexey Dosovitskiy,
\newblock ``An image is worth 16x16 words: Transformers for image recognition at scale,''
\newblock {\em arXiv preprint arXiv:2010.11929}, 2020.

\bibitem{BIT}
Hao Chen, Zipeng Qi, and Zhenwei Shi,
\newblock ``Remote sensing image change detection with transformers,''
\newblock {\em IEEE Transactions on Geoscience and Remote Sensing}, vol. 60, pp. 1--14, 2021.

\bibitem{changeformer}
Wele Gedara~Chaminda Bandara and Vishal~M Patel,
\newblock ``A transformer-based siamese network for change detection,''
\newblock in {\em IGARSS 2022-2022 IEEE International Geoscience and Remote Sensing Symposium}. IEEE, 2022, pp. 207--210.

\bibitem{review}
Ting Bai, Le~Wang, Dameng Yin, Kaimin Sun, Yepei Chen, Wenzhuo Li, and Deren Li,
\newblock ``Deep learning for change detection in remote sensing: a review,''
\newblock {\em Geo-spatial Information Science}, vol. 26, no. 3, pp. 262--288, 2023.

\bibitem{adaptSAM}
Lei Ding, Kun Zhu, Daifeng Peng, Hao Tang, Kuiwu Yang, and Lorenzo Bruzzone,
\newblock ``Adapting segment anything model for change detection in vhr remote sensing images,''
\newblock {\em IEEE Transactions on Geoscience and Remote Sensing}, 2024.

\bibitem{FASTSAM}
Xu~Zhao, Wenchao Ding, et~al.,
\newblock ``Fast segment anything,''
\newblock {\em arXiv preprint arXiv:2306.12156}, 2023.

\bibitem{SCD-SAM}
Liye Mei, Zhaoyi Ye, et~al.,
\newblock ``Scd-sam: Adapting segment anything model for semantic change detection in remote sensing imagery,''
\newblock {\em IEEE Transactions on Geoscience and Remote Sensing}, 2024.

\bibitem{MOBILESAM}
Chaoning Zhang, Dongshen Han, et~al.,
\newblock ``Faster segment anything: Towards lightweight sam for mobile applications,''
\newblock {\em arXiv preprint arXiv:2306.14289}, 2023.

\bibitem{LEVIRCC}
Chenyang Liu, Rui Zhao, Hao Chen, Zhengxia Zou, and Zhenwei Shi,
\newblock ``Remote sensing image change captioning with dual-branch transformers: A new method and a large scale dataset,''
\newblock {\em IEEE Transactions on Geoscience and Remote Sensing}, vol. 60, pp. 1--20, 2022.

\bibitem{CDCHAT}
Noman et~al.,
\newblock ``Cdchat: A large multimodal model for remote sensing change description,''
\newblock {\em arXiv preprint arXiv:2409.16261}, 2024.

\bibitem{CHANGECLIP}
Sijun Dong, Libo Wang, Bo~Du, and Xiaoliang Meng,
\newblock ``Changeclip: Remote sensing change detection with multimodal vision-language representation learning,''
\newblock {\em ISPRS Journal of Photogrammetry and Remote Sensing}, vol. 208, pp. 53--69, 2024.

\bibitem{LORA}
Edward~J Hu, Yelong Shen, Phillip Wallis, Zeyuan Allen-Zhu, Yuanzhi Li, Shean Wang, Lu~Wang, and Weizhu Chen,
\newblock ``Lora: Low-rank adaptation of large language models,''
\newblock {\em arXiv preprint arXiv:2106.09685}, 2021.

\bibitem{CLIP}
Alec Radford, Jong~Wook Kim, et~al.,
\newblock ``Learning transferable visual models from natural language supervision,''
\newblock in {\em International conference on machine learning}. PMLR, 2021, pp. 8748--8763.

\bibitem{SAM2}
Nikhila Ravi, Valentin Gabeur, et~al.,
\newblock ``Sam 2: Segment anything in images and videos,''
\newblock {\em arXiv preprint arXiv:2408.00714}, 2024.

\bibitem{FPN}
Tsung-Yi Lin, Piotr Doll{\'a}r, Ross Girshick, Kaiming He, Bharath Hariharan, and Serge Belongie,
\newblock ``Feature pyramid networks for object detection,''
\newblock in {\em Proceedings of the IEEE conference on computer vision and pattern recognition}, 2017, pp. 2117--2125.

\bibitem{iou}
Dingfu Zhou, Jin Fang, et~al.,
\newblock ``Iou loss for 2d/3d object detection,''
\newblock in {\em 2019 international conference on 3D vision (3DV)}. IEEE, 2019, pp. 85--94.

\bibitem{dice}
Carole~H Sudre, Wenqi Li, et~al.,
\newblock ``Generalised dice overlap as a deep learning loss function for highly unbalanced segmentations,''
\newblock in {\em Deep Learning in Medical Image Analysis and Multimodal Learning for Clinical Decision Support: Third International Workshop, DLMIA 2017, and 7th International Workshop, ML-CDS 2017, Held in Conjunction with MICCAI 2017, Qu{\'e}bec City, QC, Canada, September 14, Proceedings 3}. Springer, 2017, pp. 240--248.

\bibitem{LEVIR}
Hao Chen and Zhenwei Shi,
\newblock ``A spatial-temporal attention-based method and a new dataset for remote sensing image change detection,''
\newblock {\em Remote Sensing}, vol. 12, no. 10, 2020.

\bibitem{WHU}
Shunping Ji, Shiqing Wei, and Meng Lu,
\newblock ``Fully convolutional networks for multisource building extraction from an open aerial and satellite imagery data set,''
\newblock {\em IEEE Transactions on Geoscience and Remote Sensing}, vol. 57, no. 1, pp. 574--586, 2019.

\bibitem{SYSUCD}
Qian Shi, Mengxi Liu, Shengchen Li, Xiaoping Liu, Fei Wang, and Liangpei Zhang,
\newblock ``A deeply supervised attention metric-based network and an open aerial image dataset for remote sensing change detection,''
\newblock {\em IEEE Transactions on Geoscience and Remote Sensing}, vol. 60, pp. 1--16, 2022.

\bibitem{resnet}
Kaiming He, Xiangyu Zhang, Shaoqing Ren, and Jian Sun,
\newblock ``Deep residual learning for image recognition,''
\newblock in {\em Proceedings of the IEEE conference on computer vision and pattern recognition}, 2016, pp. 770--778.

\end{thebibliography}

\end{document}